\pdfoutput=1
\documentclass[11pt]{article}

\usepackage[preprint]{acl}

\usepackage{times}
\usepackage{latexsym}
\usepackage{graphicx}
\usepackage{algorithm}
\usepackage{multirow}
\usepackage{amsfonts}
\usepackage{pgfplots}
\usepackage[T1]{fontenc}
\usepackage[utf8]{inputenc}

\usepackage{microtype}

\usepackage{inconsolata}

\usepackage{graphicx}
\usepackage{subcaption}
\usepackage{algorithm}
\usepackage{algpseudocode}
\usepackage{amsmath}
\usepackage{booktabs}

\usepackage[normalem]{ulem}
\useunder{\uline}{\ul}{}

\title{LLM-Based Discriminative Reasoning for Knowledge Graph Question Answering}

\author{Mufan Xu, Kehai Chen, Xuefeng Bai, Muyun Yang, Tiejun Zhao, Min Zhang\\
  School of Computer Science and Technology, Harbin Institute of Technology, China \\
  \texttt{xmuffins0610@gmail.com},\\  \texttt{\{chenkehai,baixuefeng,yangmuyun,tjzhao,zhangmin2021\}@hit.edu.cn}}

\begin{document}
\maketitle
\begin{abstract}
%In knowledge graph question answer, knowledge graph reasoning includes the subgraph retrieval and the answer inference.
Large language models (LLMs) based on generative pre-trained Transformer have achieved remarkable performance on knowledge graph question-answering (KGQA) tasks. 
However, LLMs often produce ungrounded subgraph planning or reasoning results in KGQA due to the hallucinatory behavior brought by the generative paradigm.
%, which may hinder the advancement of the LLM-based KGQA method. 
To tackle this issue, we propose READS to reformulate the KGQA process into discriminative subtasks, which simplifies the search space for each subtasks.
Based on the subtasks, we design a new corresponding discriminative inference strategy to conduct the reasoning for KGQA, thereby alleviating hallucination and ungrounded reasoning issues in LLMs.
% 说特点，用什么方法解决了什么问题，可以截上面一句的后半段
Experimental results show that the proposed approach outperforms multiple strong comparison methods, along with achieving state-of-the-art performance on widely used benchmarks WebQSP and CWQ.~\footnote{\textcolor{blue}{Our code and data will be released upon acceptance.}}
% available at \url{https://github.com/anonymous}.

% Knowledge graph reasoning based on large language models (LLMs) has achieved remarkable performance on KGQA tasks through direct planning or interactive search. However, due to LLMs' hallucinatory behavior in the planning and reasoning processes, previous proposed methods may produce ungrounded path planning or reasoning results. In this paper, we propose Graph Reasoning with Structural Search (READS) to reformulate the KGQA task as a discriminative graph reasoning process based on LLMs. By utilizing graph structure-based discriminative strategies, READS enhances the LLMs' capability to precisely retrieve question-related knowledge subgraph and perform reliable reasoning on it. Experiments show that READS alleviates hallucination problems during KGQA process and achieves state-of-the-art performance on two widely used benchmarks WebQSP and CWQ, along with significant improvements in the reliability of the complex graph reasoning process based on LLMs.

\end{abstract}

\section{Introduction}
\label{sec:intro}

% 研究对象的好，和第二段对应上

Large language models (LLMs) have shown remarkable reasoning capabilities in KGQA task~\cite{yudecaf,huang2023towards,zhu2024benchmarking}, especially the feasibility to prompt the LLMs to generate searching and reasoning results through the LLMs' built-in knowledge. 
Typically, based on the given question, LLMs can be prompted to provide a plan for the question-related subgraph through one-time generation. 
After retrieving the subgraph, LLMs can directly generate the answers along with the reasoning steps using the subgraph as context.
Utilizing internal knowledge or reasoning ability distilled from stronger models like GPT-4, the generative KGQA model can effectively conduct knowledge graph reasoning, along with achieving state-of-the-art performance on the KGQA tasks~\cite{mondorf2024beyond,rog,ToG}.

%改一下
\begin{figure}[htbp]
    \centering
    \includegraphics[width=0.87\linewidth]{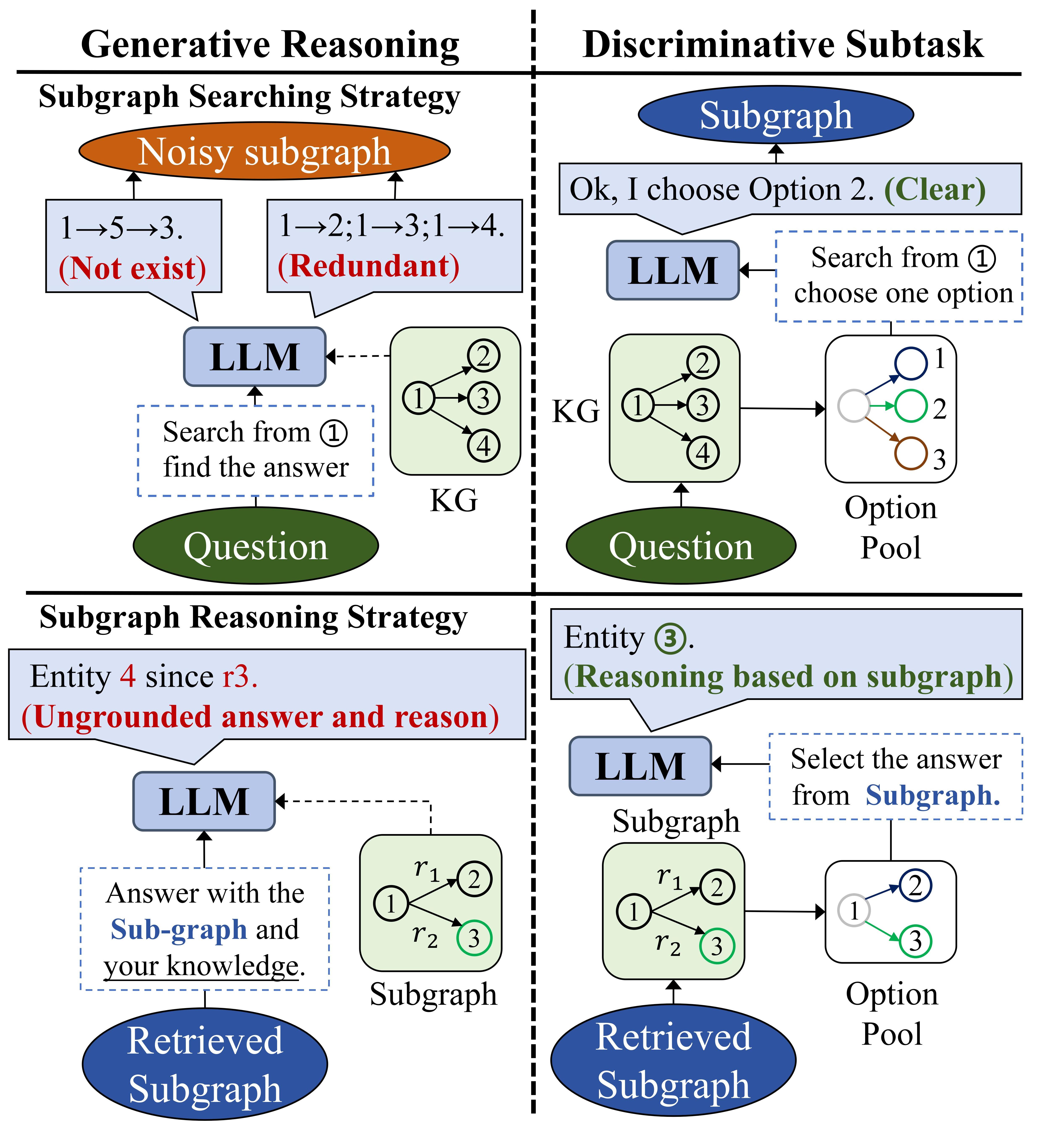}
    \caption{The generation-based methods tend to generate unsupported or redundant subgraphs and reasoning results (left), while the proposed method address the issue by establishing proper searching space for each of the KGQA subtasks (right).}
    \label{fig:fig2}
\end{figure}

% Representative methods like RoG first generate target paths (\citealp{wang2023plan}) and then conduct chain-of-thought reasoning to reach the answer (\citealp{rog}), while ToG turn to iteratively expands the subgraph and then prompts the LLM to directly answer the question (\citealp{ToG}). 

Despite their success, the generative reasoning methods often produce ungrounded planning or reasoning results due to the hallucinatory behavior~\cite{zhang2023siren,sun2024head,pan2024unifying}, which is opposite to the deterministic characteristic of knowledge reasoning process~\cite{garcez2015neural,xiong2024converging}. 
As shown in Figure~\ref{fig:fig2}, when searching question-related subgraphs, generation-based methods come with not existed path ``\textit{1->5->3}” due to hallucinatory planning , or retrieve redundant paths at one step ``\textit{1->2; 1->3; 1->4}” as a compensate to the generation uncertainty (upper left). 
% These issues lead to a redundant retrieved subgraph, which disperses the LLM's attention and may trigger hallucinatory reasoning results. 
When conducting answer inference on the retrieved subgraph, the generation-based methods may generate unreasonable step ``\textit{since r3}" as inference chain or even entity ``\textit{4}” that do not exist in the subgraphs as answers (bottom left).
The hallucinatory behavior of the generative LLMs hinders the advancement of KGQA. 
%Adopting discriminative strategies may alleviate the issue brought by generation-based methods and achieve subgraph searching and subgraph reasoning results accord with the knowledge graph (KG) and retrieved subgraph (upper and bottom right).

To address the issue, we propose LLM-Based \textbf{Rea}soning With \textbf{D}iscriminative \textbf{S}ubtasks (READS) to strengthen the LLM-based knowledge reasoning process.
%两三句话 技术路线 特点
READS decomposes the KGQA process into three discriminative subtasks: graph searching, graph pruning, and answer inference. The decomposition aims to explicitly simulate the capabilities of searching for question-related knowledge, identifying semantic constraints, and inferring the answer position on the subgraph, respectively.
Meanwhile, READS simplifies search space from the knowledge graph without toolboxes, along with designed discriminative inference strategy to conduct the reasoning of KBQA effectively.
In summary, our main contributions are as follows:
%We conducted extensive experiments to support our claims and further verify the advantages of READS on two widely used KGQA benchmarks WebQSP (\citealp{webqsp}) and CWQ (\citealp{cwq}). In general, the main contributions of this work are:

%第二个贡献应该是提出方法的特点，高效推理
\begin{itemize}
    \item We introduce READS, an novel reasoning framework that explicitly models KGQA reasoning skills by deconstructing the KGQA process into three discriminative subtasks.
%reduce the hallucinatory behavior brought by generation-based KGQA strategies.
    \item An effective corresponding discriminative inference strategy is designed to conduct the reasoning of KGQA for READS, thereby significantly alleviating hallucination and ungrounded reasoning issues.
    %By the discriminative reasoning framework, the proposed READS method not only enhances the capability of LLM to retrieve question-related subgraphs, but also alleviates the issue of ungrounded reasoning brought by the LLM generation process.
    \item Experimental results demonstrate that READS achieved state-of-the-art performance on two widely used benchmarks.
\end{itemize}

\section{Related Works}

%总括一下每段内容
\textbf{Generative Approaches.} The challenge of KGQA task lies in how to conduct precise reasoning on the knowledge graphs~\cite{miller-etal-2016-key,bai-etal-2022-graph,zhang-etal-2024-question}, early works tried to teach models to construct database queries for knowledge graphs, allowing them to directly retrieve answers from the graph~\cite{gu-su-2022-arcaneqa,ye-etal-2022-rng}. With the advent of LLM's long-horizon planning and reasoning capability~\cite{zhong2024memorybank, wang2024survey}, the focus of KGQA research shifts toward leveraging the reasoning capabilities of a single LLM for knowledge inference~\cite{jiang2022unikgqa}. One straightforward way is to directly schedule the question-related sub-graph using the LLM's knowledge~\cite{hong2023faithful,wang2023knowledgedrivencotexploringfaithful}. Typical approach like RoG employs chain-like subgraph planning and distills GPT-4’s Chain-of-Thought reasoning capability to achieve reliable reasoning processes over knowledge graphs, achieving state-of-the-art performance \citep{rog}. Despite their success, one concern is that those methods often provides incorrect and ungrounded reasoning results due to the hallucinatory generation process.

% 提出直接生成方法的问题，引出interactive类型的方法，审稿人提出的近期工作GoG是ToG类型的延申，多了每一步之前的一个动机生成来引导当次决策进行的行动内容，当然还是通过生成和prompting方法，基于GPT-4；因而和ToG一样仍然受幻觉行为影响。并且这些方法由于允许GPT-4等强模型运用自己的知识，模型对知识的错误记忆成为新的幻觉行为因素，使得对推理过程的分析复杂化。

\noindent\textbf{Interactive and Discriminative Approaches.} An alternative approach is to design effective tools and generation strategies to retrieve environment information from the knowledge graph to enhance the step-by-step reasoning process, as seen in approaches such as ToG, KGAgent and GoG~\cite{ToG,jiang2024kg,xu2024generate}. However, these interactive generation approaches cannot avoid the influence of hallucinations. Even when provided with environmental information or recalled subgraphs, the model may still arrive at incorrect reasoning results. As claimed in PANGU, using discriminative strategy can effectively mitigates the hallucination problem~\cite{gu2023don}. Despite PANGU's success, integrating tools such as search and answer retrieval within the same search space may also lead the LLM to make erroneous decisions.

% This issue becomes even more difficult to analyze when powerful models like GPT-4 are allowed to use their internal parameterized knowledge, as reasoning errors can be obscured by the model's incorrect memory of knowledge.

%引出pangu，判别式方法是缓解幻觉行为，提升KGQA稳定性的方法。这里要说明我们的方法与PANGU之间的区别，即我们沿用了判别式的思路，但是进一步对混杂的任务目标进行了拆分，实现了混杂任务的语义解耦，进一步提升了过程的可靠性。

In this paper, we propose to reformulate KGQA into three subtasks to explicitly model the KGQA skills and design discriminative strategies to effectively enhance LLM's reasoning capability on the knowledge graph.

% In this paper, we extend the discriminative approach by decoupling the decision-making tasks into search, constraint, and answering through task reformulation. This approach highlights the model's ability to cognitively process the sub-graph structure pertinent to the query, thereby facilitating a more stable KGQA process.

%框架
\section{READS Framework}
\label{sec:back}

\begin{figure*}
    \centering
    \includegraphics[width=1\linewidth]{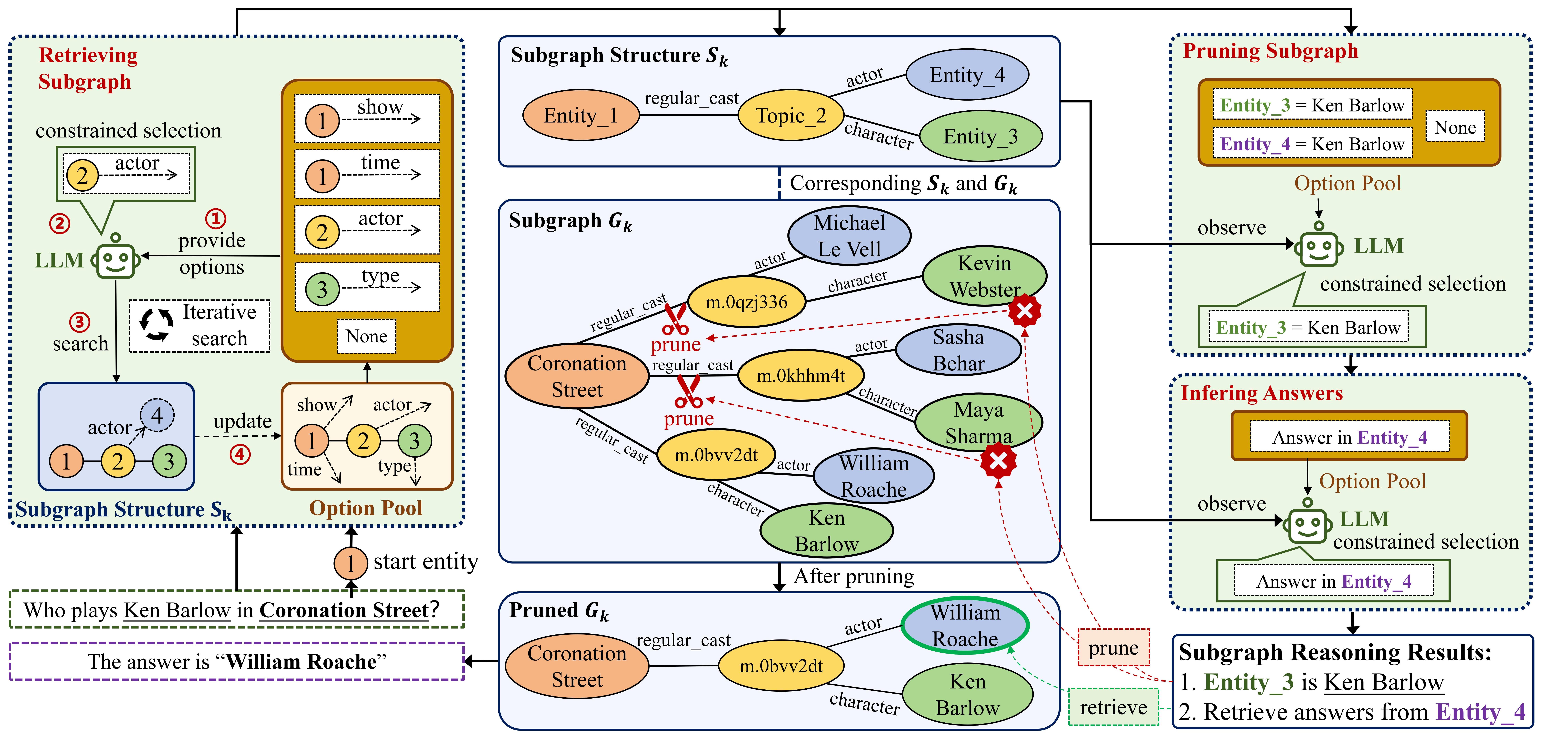}
    \caption{The proposed READS for KGQA. Start from the question (bottom left) with a given starting entity "Coronation Street", READS sequentially conducts subgraph retrieval, subgraph pruning, and answer inference. Then READS automatically uses the reasoning results to prune $G_k$ and then retrieve the answers from it. In this figure, the node's color in subgraph $G_k$ (middle center) represents its position in subgraph structure $S_k$ (top center).}
    \label{fig:fig1}
\end{figure*}

% When applying the subgraph reasoning results (bottom right), READS prune the neighboring edges of start entity in $G_k$, and then retrieve the answers corresponding to the selected answer node (Entity\_4 in blue).

In this section, we propose a novel framework that reformulates the KGQA task into three discriminative subtasks, including question-related subgraph searching, question-related subgraph pruning, and answer inference. %Here we first formulate the KGQA task.

\noindent\textbf{Formulation of KGQA task.} Given a question $Q$ and the knowledge graph entities $E$ contained in the question, the KGQA task asks the model to recall golden answers $A_{gold}$ as much as possible. The model has to retrieve question-related subgraphs from the knowledge graph and infer the right answer based on the subgraphs.

%\subsection{Knowledge Graph Question Answering}
The knowledge graph $KG$ used in this paper is Freebase\footnote{The two benchmarks used in this work are constructed using Freebase \cite{bollacker2008freebase}.}, which consists of knowledge triplets represented as $t=(s,p,o)$ including the subject entity $s$, the object entity $o$, and the predicate $p$ that connects these two entities.

\subsection{Question-related subgraph Searching}
\label{sec:concept}
Given the knowledge graph, the question-related subgraph searching subtask aims to retrieve the question-related subgraph $G_{k}$ and thereby summarizes an abstract structure $S_k$ for each input question. 
Specifically, $G_{k}$ is the subgraph comprising only the necessary knowledge to correctly answer the question $Q$, which can be represented by a set of triplets:
\begin{equation}
\begin{aligned}
G_{k}=\{t_i|t_i=(s_i,p_i,o_i)\}.
\end{aligned}
\end{equation}
Following UniKGQA~\cite{jiang2022unikgqa}, we use ``semantic nodes'' to represent a group of entities sharing same structural position in the knowledge graph. The summarized abstract structure $S_{k}$ groups all the entities into ``semantic nodes” based on their position in $G_{k}$:
\begin{equation}
\begin{aligned}
S_{k}=\{t_i|t_i=(s_{abs},p_j,o_{abs})\},\\
s_{abs},o_{abs}\in Group(G_k).
\end{aligned}
\end{equation}
For example, if $G_{k}$ includes two triples, ($a$, $friend\_of$, $b$) and ($a$, $friend\_of$, $c$), its abstract structure $S_k$ is $\{entity_1, friend\_of, entity_2\}.$
Here both entities $b$ and $c$ are grouped into the abstract node $entity\_2$ since they connect to the same entity $a$ with the same relation $friend\_of$. Note that $S_k$ only groups the entities and keeps the name of the relations.

% Meanwhile, question-related subgraph $G_{k}$ can also be decomposed into a set of sub-trees $T_{n}$ that satisfy $S_{k}$:
% $$G_{k} = T_{1}+T_{2},$$
% $$T_{1} = \{(a, friend\_of, b)\},$$
% $$T_{2} = \{(a, friend\_of, c)\}.$$
% Assume that the $G_k$ above is the subgraph retrieved from Freebase for the question \textit{"who is a's friend?"}, then each one of $T_{1}$ and $T_{2}$ can provide sufficient evidence for a candidate answer.

\subsection{Question-related Subgraph Pruning}
Based on the abstract structure $S_{k}$, the question-related subgraph pruning task aims to map all the question-related constraint entities $C$ to nodes in $S_{k}$.
$C$ denotes the intersection of question mentioned entities $E_{question}$ and all entities in Freebase $E_{freebase}$: 
\begin{equation}
C = E_{question} \cap E_{freebase}.
\end{equation}
Let Node($S_k$) represents the set of nodes in $S_k$, the mapping results between entities in $C$ and nodes in $S_k$ is represented as:
\begin{equation}
\{(C_i,N_i)|C_i \in C, N_i \in Node(S_{k})\}.
\end{equation}
For example, when answering the question: \textit{"What is the name of the team who won the Super Bowl in 2011?"}, there are two constraint entities $C_{1}$: "Super Bowl" and $C_{2}$: "2011". Now that the retrieved subgraph $G_k$ is always a tree rooted from the starting entity, any branches contain information against the the information in $C_{1}$ should be pruned from its root. In order to better focus on the LLM-based discriminative reasoning process, we assume that the entities mentioned in the questions have already been linked to Freebase entities through rule-based recognition methods.

\subsection{Answer Inference}

Given the question-related subgraph structure $S_{k}$, the answer inference subtask aims to locate the position of the answer $A_{pos}$ which corresponds to the position of $A_{gold}$ in $S_k$:
\begin{equation}
\begin{aligned}
A_{pos} = \text{Grouped}(A_{gold}),\\
A_{gold} \in \text{Node}(G_k),\\
A_{pos} \in \text{Node}(S_k).
\end{aligned}
\end{equation}
Once the position of the answer $A_{pos}$ is selected, the corresponding group of entities in $G_k$ will be regarded as the final answers.

%为什么会有三个策略 没交代
\section{READS Discriminative Reasoning}
\label{sec:approch}
After we propose the framework of READS, we are able to design efficient reasoning strategy to facilitate graph retrieval, graph pruning and answer inference. 
Based on the subtasks, we are able to explicitly model the KGQA process as illustrated in Figure~\ref{fig:fig1}. We design discriminative strategies to achieve the subtasks and construct training data to augment the LLM-based reasoning process.

% To better integrate structural information into the reasoning process, we propose a two stage method READS: 1) \textbf{"tree search" module} that performs iterative tree search and 2) \textbf{"tree pruning" module} that performs structural pruning. READS directly lists all knowledge graph nodes situated on the chosen structural position as the answer. 
% Different from RoG and ToG~\cite{rog,ToG},  To map the problem to its corresponding knowledge subgraph structure $S_k$, READS interacts with knowledge graph and expands the retrieved knowledge subgraph structure $S_{k}$ from scratch.

\subsection{Searching Strategy}

Compared to previous methods that rely on agent toolboxes, READS opts for a discriminative searching approach which only observes and updates the subgraph structure $S_k$. 

In each iteration, based on the retrieved subgraph structure $S_k$, the LLM selects one option from the option pool (as shown in Figure~\ref{fig:fig1}). Each option includes a starting node $s_{next}$ in $S_k$ and a neighboring relation $p_{next}$, forming the next triple $t_{next}=(s_{next},p_{next},o_{next})$. A new node $o_{next}$ is added to $S_k$ along with its corresponding entities in $G_k$ retrieved from the knowledge graph. READS maintains an option pool that includes all feasible triples for search. We can formulate each step of the discriminative searching strategy in READS as:
\begin{equation}
    t_{next}=argmax(\mathbb{P}(t|S_{k},Q),t\in pool),
\end{equation}
where $\mathbb{P}$ represents the probability distribution over options provided by the LLM using a constrained beam search algorithm based on output logits. An additional option, `None', is always available to terminate the search process.

\noindent\textbf{Retention of Node Information.} READS further enriches the information contained within the subgraph structure $S_{k}$ by labeling the semantic nodes with entity types. Entities in Freebase can be classified into one of the following types: entity, topic, date, and num (details are provided in Appendix~\ref{sec:appendix_A}). READS will recognize the type of retrieved entities, and add new nodes in $S_{k}$.

\subsection{Pruning Strategy}
\label{sec:prune}

After obtaining the question-related subgraph $G_k$ along with its structure $S_k$, READS maps all constraints mentioned in the question onto $S_k$ to perform subgraph pruning. A triplet $C_{n}$ = ($c_{pos}$, $c_{opt}$, $c_{tar}$) has to be chosen from the option pool, where $c_{tar}$ is the constraint entity mentioned in the question, $c_{pos}$ is the target position for applying the constraint, and $c_{opt}$ is the operator to define the type of logical resolution used for applying the constraint. READS restrict the operator $c_{opt}$ to one of the seven types $\{=, <, \le, >, \ge, \text{min}, \text{max}\}$ and combine each constraint with all possible operators and positions to form the option pool. 

Based on the question $Q$ and $S_{k}$, READS asks the LLM to iteratively select the constraints until the LLM selects `None' or there are no options left:
\begin{equation}
    C_{n} = argmax(\mathbb{P}(C_{n}|C_{1}...,C_{n-1},S_{k},Q)).
\end{equation}

The pruning process is conducted at the level of subtrees rooted from the starting node in $G_k$ (as shown in Figure~\ref{fig:fig1}), retaining only the subtrees that meet the constraints.

\begin{table*}[]
\centering
\scalebox{0.85}{
\begin{tabular}{llllllll}
\toprule[1pt]
\multirow{2}{*}{\textbf{Method}} &  \multicolumn{3}{c}{\textbf{WebQSP}} & \multicolumn{3}{c}{\textbf{CWQ}} \\ \cline{2-7} 
 &Hits@1       &Recall       &F1      &Hits@1       &Recall       &F1 \\ \hline
Llama2-7b zero-shot (\citealp{touvron2023llama})* &0.403       &-       &0.293      &0.297       &-       &0.272      \\
Llama3-8b zero-shot (\citealp{dubey2024llama})* &0.303       &-       &0.257      &0.305       &-       &0.278      \\
Qwen2.5-7b zero-shot (\citealp{yang2024qwen2})* &0.284       &-       &0.237      &0.259       &-       &0.241      \\
GPT-4-turbo zero-shot~\cite{achiam2023gpt}* &0.632       &-       &-      &0.483       &-       &-      \\
Llama2-7b SPARQL Generation* &0.747       &-       &-      &0.656       &-       &-      \\ \hline
KV-Mem (\citealp{miller-etal-2016-key}) &0.467       &-       &0.345      &0.184       &-       &0.157      \\
GraftNet (\citealp{sun-etal-2018-open}) &0.664       &-       &0.604      &0.368       &-       &0.327      \\
QGG (\citealp{lan-jiang-2020-query}) &0.730       &-       &0.738      &0.369       &-       &0.374      \\
NSM (\citealp{He_2021}) &0.687       &-       &0.628      &0.476       &-       &0.424      \\
SR+NSM+E2E (\citealp{zhang-etal-2022-subgraph}) &0.695       &-       &0.641      &0.493       &-       &0.463      \\
DECAF (DPR+FiD-3B) (\citealp{yudecaf}) &0.821       &-       &0.788      &-       &-       &-      \\
UniKGQA (\citealp{jiang2022unikgqa}) &0.772       &-       &0.722      &0.512       &-       &0.490      \\
PANGU (\citealp{gu2023don}) &0.796       &-       &-      &0.622       &-       &-      \\
KD-CoT (\citealp{wang2023knowledgedrivencotexploringfaithful}) &0.686       &-       &0.525      &0.557       &-       &-      \\
ToG w/GPT-4 (\citealp{ToG}) &0.826       &-       &-      &0.676       &-       &-      \\
KG-Agent (\citealp{jiang2024kg}) &0.833       &-       &0.810      &0.722       &-       &0.692      \\
RoG (Top-3 relation path) (\citealp{rog})* &0.795       &0.756       &0.701      &0.567       &0.573       &0.547      \\ \hline
READS (Ours)&\textbf{0.840}       &\textbf{0.860}     &\textbf{0.845}      &\textbf{0.802}       &\textbf{0.837}       &\textbf{0.820}      \\ \bottomrule[1pt]
\end{tabular}
}
\caption{The results of our method compared with previous approaches on WebQSP and CWQ. Asterisk (*) denotes the results we reproduced. Note that the Hits@1 result reported in the original RoG paper (WebQSP 0.857, CWQ 0.626) is not calculated in the right way, see the author's response \href{https://github.com/RManLuo/reasoning-on-graphs/issues/11}{here}.}
\label{table:main}
\end{table*}

\subsection{Answering Strategy}

When answering questions with a large number of answers, previous generative methods often fail to capture all the correct answers, even if the reasoning steps are successfully generated. To address this problem, READS focuses on locating the positions of answers within the subgraph structure $S_k$ to simultaneously retrieve all possible answers. Based on $S_{k}$, READS determines the answer position $A_{pos}$ using:
\begin{equation}
    A_{pos} = argmax(\mathbb{P}(n|S_k,C,Q),n\in S_k),
\end{equation}
where $\mathbb{P}$ is also given by the LLM based on $S_k$,$C$, and $Q$. Positions for applying constraints can not be chosen again. All entities in $G_k$ corresponding to the position $A_{pos}$ will be listed as the answer. 

\section{Experiments}
\label{sec:result}

\subsection{Datasets and Settings}

\textbf{Data preprocessing.} Based on the proposed subtasks, we construct training data based on the original training sets. We get 121,023 subgraph searching samples and 46,885 subgraph pruning and answer inference samples. For more details of our data preprocessing and training data construction method, please refer to Appendix~\ref{sec:appendix_B}.

\noindent\textbf{Benchmarks.} To evaluate the knowledge graph question-answering capability of the proposed method, we choose two widely used benchmarks, WebQSP (\citealp{webqsp}) and CWQ (\citealp{cwq}). These two benchmarks are constructed based on Freebase knowledge graph.

\noindent\textbf{Metrics.} We choose commonly used metrics Hits@1 and F1 for the evaluation process following previous works~\cite{rog,ToG}. For detailed definition and implementation of the metrics, please refer to Appendix~\ref{sec:appendix_C}.

\noindent\textbf{Baselines} We use previous reproducible SOTA generation-based KGQA method RoG as our baseline. RoG make full use of LLM planning and chain-of-thought reasoning capability to achieve remarkable KGQA performance~\cite{rog}. We also listed typical methods like ToG and KGAgent with interactive reasoning strategy (\citealp{ToG,jiang2024kg}), PANGU with single-task discriminative strategy~\cite{gu2023don}. The zero-shot performance of widely used LLMs is listed for comparison. We also finetuned llama2-7b to directly generate SPARQL queries for each of the question, and then execute those queries on Freebase to get the answer. 

\noindent\textbf{Base Model.} We choose Llama2-7b as the base model of READS following RoG. For implementation with GPT-4, see Section~\ref{sec:gpt4}.

\subsection{Main results}

The performance of READS on WebQSP and CWQ is presented in Table~\ref{table:main}. According to the results, our porposed READS framework achieves state-of-the-art performance on these two benchmarks, with improvements in both Hits@1 and F1, indicating an enhanced capability of the LLM to handle KGQA tasks. Besides, the READS method abandons the use of internal model knowledge, yet still achieves better KGQA performance, which sufficiently demonstrates that the proposed framework can effectively enhance the knowledge reasoning capabilities of LLMs (refer to Appendix~\ref{sec:appendix_D}). We also test READS on more challenging dataset GrailQA~\cite{gu2021beyond}, the results are shown in Appendix~\ref{sec:grailqa}.

% \subsection{Measuring the Quality of Retrieved subgraphs}

% \section{Discussion}

% We have demonstrated the performance improvements brought by our method. To further validate the effectiveness of our method and support our claims, we conduct a series of experiments.

\subsection{Searching Capability Analysis}
\label{subsec:qualityofgraph}

\begin{table*}[h]
\centering
\scalebox{0.71}{
\renewcommand{\arraystretch}{1.3}
\begin{tabular}{lccccccccccccccc}
\toprule[1pt]
\multicolumn{1}{l|}{\textbf{Leaf Number}} & \multicolumn{5}{c|}{\textbf{2}}                                                                         & \multicolumn{3}{c|}{\textbf{3}}                                       & \multicolumn{3}{c|}{\textbf{4}}                                       & \multicolumn{3}{c|}{\textbf{5}}                                       & \multirow{2}{*}{\textbf{Avg.}} \\ \cline{1-15}
\multicolumn{1}{l|}{\textbf{Total Hop}}   & \textbf{1}     & \textbf{2}     & \textbf{3}     & \textbf{4}     & \multicolumn{1}{c|}{\textbf{5}}     & \textbf{3}     & \textbf{4}     & \multicolumn{1}{c|}{\textbf{5}}     & \textbf{4}     & \textbf{5}     & \multicolumn{1}{c|}{\textbf{6}}     & \textbf{5}     & \textbf{6}     & \multicolumn{1}{c|}{\textbf{7}}     &                                \\ \hline
\multicolumn{16}{l}{\textbf{Relation Recall $R_{rel}$}}                                                                                                                                                                                                                                                                                                                                                                \\ \hline
\multicolumn{1}{l|}{RoG}                  & 0.853          & 0.644          & 0.381          & 0.280          & \multicolumn{1}{c|}{0.254}          & 0.429          & 0.266          & \multicolumn{1}{c|}{0.270}          & 0.286          & 0.179          & \multicolumn{1}{c|}{0.169}          & 0.186          & 0.266          & \multicolumn{1}{c|}{0.283}          & 0.339                          \\
\multicolumn{1}{l|}{READS}          & \textbf{0.887} & \textbf{0.887} & \textbf{0.903} & \textbf{0.897} & \multicolumn{1}{c|}{\textbf{0.972}} & \textbf{0.859} & \textbf{0.853} & \multicolumn{1}{c|}{\textbf{0.899}} & \textbf{0.748} & \textbf{0.656} & \multicolumn{1}{c|}{\textbf{0.826}} & \textbf{0.867} & \textbf{0.837} & \multicolumn{1}{c|}{\textbf{0.863}} & \textbf{0.853}                 \\ \hline
\multicolumn{16}{l}{\textbf{Minimum Graph Edit Distance $D(G_{k},G_{gold})$}}                                                                                                                                                                                                                                                                                                                                                            \\ \hline
\multicolumn{1}{l|}{RoG}                  & 0.479          & 2.494          & 3.929          & 5.462          & \multicolumn{1}{c|}{7.727}          & 3.071          & 3.746          & \multicolumn{1}{c|}{5.394}          & 1.760          & 4.780          & \multicolumn{1}{c|}{5.681}          & 5.441          & 8.100          & \multicolumn{1}{c|}{10.438}         & 4.893                          \\
\multicolumn{1}{l|}{READS}          & \textbf{0.097} & \textbf{0.209} & \textbf{0.315} & \textbf{0.625} & \multicolumn{1}{c|}{\textbf{0.181}} & \textbf{0.338} & \textbf{1.069} & \multicolumn{1}{c|}{\textbf{1.490}} & \textbf{1.521} & \textbf{3.658} & \multicolumn{1}{c|}{\textbf{3.000}} & \textbf{1.235} & \textbf{1.550} & \multicolumn{1}{c|}{\textbf{1.578}} & \textbf{1.204}                 \\ \bottomrule[1pt]
\end{tabular}
}
\caption{We use relation recall and minimum graph edit distance as the metrics to measure the quality of retrieved subgraphs with different type of structures.}
\label{table2}
\end{table*}

To validate READS's enhancement on the capability of LLM to retrieve question-related subgraphs, we design two metrics, relation recall and minimum graph edit distance, to measure the difference between the retrieved subgraph $G_{k}$ and the golden subgraph $G_{gold}$ extract from the SPARQL query given by the benchmarks.

Relation recall measures the proportion of golden relations edges that are successfully predicted, which reflects the method's sensitivity to retrieve the most relevant relations towards the given question from the knowledge graph:

{\small
\begin{equation}
R_{rel} = \frac{\text{count}(\{R|R\in G_{k}\}\cap \{R|R\in G_{gold} \})}{\text{count}(\{R|R\in G_{gold}\})}.
\end{equation}
}

Minimum edit distance $D(G_{1},G_{2})$ is defined as the total number of operations required to transform one graph $G_{1}$ into another graph $G_{2}$ by sequentially adjusting its edges one by one:
{
\begin{equation}
D(G_{1},G_{2}) = \underset{n}{min} (\text{Edit}^{n}(G_{1})==G_{2}).
\end{equation}
}

The lower the distance $D(G_{k},G_{gold})$ is, the smaller the structural difference between the predicted graph structure and the correct reasoning subgraph is. We combine the WebQSP and CWQ datasets and classify the test set based on the structure of the given golden subgraph with two features: number of leaf nodes and the total number of relations. The detailed statistic result can be found in Appendix~\ref{appendix:sta_subgraph}. We compare READS with finetuned generative method RoG to evaluate the method's performance to retrieve different types of subgraphs, the results are shown in Table~\ref{table2}.

Across all types of subgraph structures, it is evident that our method consistently achieves higher relation recall and lower average edit distance, which demonstrates significant enhancement of the LLM's capability to search for question-related subgraphs with our proposed searching strategy.

\subsection{Pruning-Answering Capability Analysis}
\label{sec:stability}

Following the analysis of searching capability, we move on to evaluate the pruning and answering capability of our proposed strategies. We calculate the average size of the retrieved subgraphs. As shown in Figure~\ref{fig:boxplot}, there is a significant reduction in the average size of the retrieved subgraphs, indicating that the READS method effectively improves the efficiency of subgraph retrieval by recalling fewer but higher-quality subgraph triples.

\begin{figure}[h]
    \centering
    \begin{subfigure}[b]{0.47\columnwidth}
        \centering
        \begin{tikzpicture}
            \begin{axis}[
                width=4.6cm,  
                height=4.6cm, 
                bar width=4pt,
                ymode=log,
                ybar,
                ybar interval=0.7,
                enlargelimits = 0.02,
                legend pos=north east,
                legend style = {font=\scriptsize},
                ymajorgrids=true,
                xmajorgrids=false,
                symbolic x coords={20, 60, 100, 140, 180, 200+},
                clip mode=individual,
                xlabel={size of retrieved subgraphs},
                tick label style={font=\tiny},
                xlabel style={font=\scriptsize,yshift=0pt},
            ]

            \addplot coordinates {(20, 1175) (60, 195) (100, 69) (140, 36) (180, 37) (200+, 30)};
            \addplot coordinates {(20, 1455) (60, 61) (100, 46) (140, 14) (180, 20) (200+, 24)};
            \legend{RoG, READS}
            \end{axis}
        \end{tikzpicture}
        \caption{WebQSP}
        \label{fig:sub-first1}
    \end{subfigure}
    \hspace{0.02\columnwidth}
    \begin{subfigure}[b]{0.47\columnwidth}
        \centering
        \begin{tikzpicture}
            \begin{axis}[
                width=4.6cm,  
                height=4.6cm, 
                bar width=4pt,
                ymode=log,
                ybar,
                ybar interval=0.7,
                enlargelimits = 0.02,
                legend pos=north east,
                legend style = {font=\scriptsize},
                ymajorgrids=true,
                xmajorgrids=false,
                symbolic x coords={20, 60, 100, 140, 180, 200+},
                clip mode=individual,
                xlabel={size of retrieved subgraphs},
                tick label style={font=\tiny},
                xlabel style={font=\scriptsize,yshift=0pt}, 
            ]

                \addplot coordinates {(20, 1893) (60, 397) (100, 177) (140, 120) (180, 158) (200+, 208)};
                \addplot coordinates {(20, 2765) (60, 249) (100, 71) (140, 76) (180, 22) (200+, 264)};
            \legend{RoG, READS}
            \end{axis}
        \end{tikzpicture}
        \caption{CWQ}
        \label{fig:sub-first2}
    \end{subfigure}
    \caption{The number of cases with the size (number of triplets) of retrieved subgraphs.}
    \label{fig:boxplot}
\end{figure}

\begin{figure}[h]
    \centering
    \includegraphics[width=0.85\linewidth]{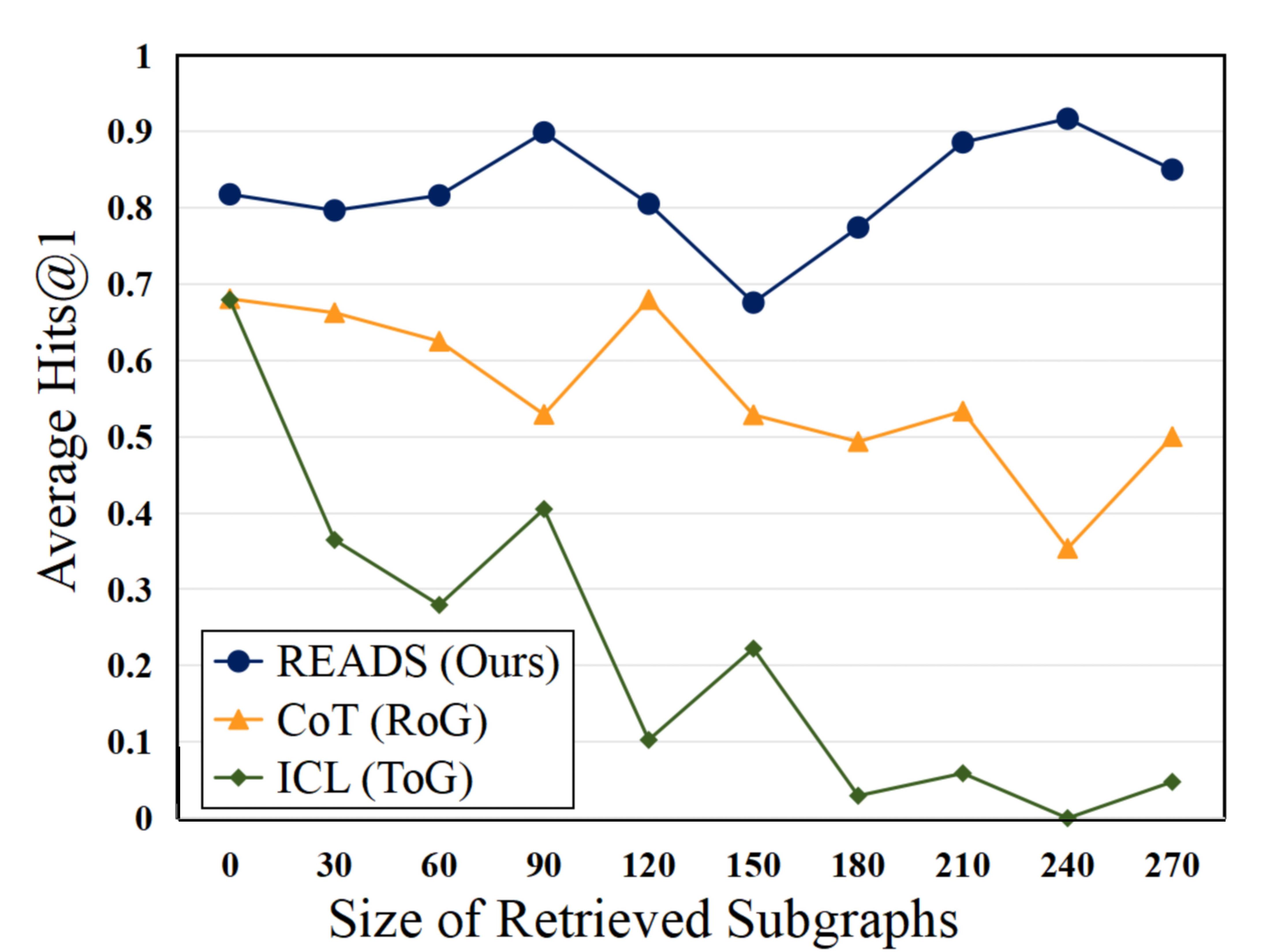}
    \caption{The trend of average Hits@1 as the size (number of triplets) of retrieved subgraph increases.}
    \label{fig:fig4}
\end{figure}

% \begin{figure}
%     \centering
%     \includegraphics[width=0.8\linewidth]{figure5.jpg}
%     \caption{The average size (number of triplets) of retrieved subgraphs.}
%     \label{fig:boxplot}
% \end{figure}

We analyze the impact of the size of the retrieved subgraph (i.e., the number of triples included) on the overall performance of the strategy (the result is shown in Figure~\ref{fig:fig4}). In addition to using RoG with finetuned chain-of-thought reasoning, we implement the in-context reasoning strategy proposed by ToG with the subgraphs retrieved by READS. To ensure fairness, we use Llama2\_7b as the base model for all experiments. As shown in Figure~\ref{fig:fig4}, as the number of recalled subgraph triples increases, the performance of generative reasoning methods declines, whereas the strategy adopted by READS remains more stable. Redundant subgraph significantly increase the context length, thereby affecting the performance of the generative reasoning process. This observation suggests that generation-based reasoning strategies are more sensitive to the size of the subgraphs compared to the strategy employed by READS.

\subsection{Subtask Ablation Study}

To validate the effectiveness of the reformulation approach adopted in READS, we ablate the strategies in READS one at a time and observe the changes in performance. The implementations are: 1) answer: Generate the answer based on the subgraph $G_k$ rather than determining position on $S_k$;  2) pruning: Skip the pruning process and rely on answer generation process to filter answers; 3) searching: Directly generate subgraph paths based on the question using the strategy in RoG. 4) entity type: Erase the entity type on $S_k$, an extra ablation implementation to evaluate the effectiveness of entity information retention for LLM reasoning.

\begin{table}[h]
\scalebox{0.9}{
\renewcommand{\arraystretch}{1}
\centering
\begin{tabular}{l|cc|cc}
\hline
\multirow{2}{*}{\textbf{Model}}          & \multicolumn{2}{c|}{\textbf{WebQSP}}   & \multicolumn{2}{c}{\textbf{CWQ}}       \\ \cline{2-5}
                                 & Hits@1 & F1    & Hits@1 & F1    \\ \hline
READS               & 0.840  & 0.845 & 0.802  & 0.820 \\ 
- answer & 0.761  & 0.744 &0.684        &0.679       \\
                              
- pruning             & 0.737  & 0.764 &0.548        &0.632       \\
- searching               & 0.739  & 0.803 & 0.444  & 0.581  \\
                                  \hline
- entity type            & 0.764  & 0.776       & 0.741  & 0.770        \\ \hline 
\end{tabular}
}
\caption{Ablation study of the strategies in READS.}
\label{strategy_ablation}
\end{table}

The results are shown in Table~\ref{strategy_ablation}. Firstly, all three tasks experienced a performance decline when employing strategies similar to previous work, demonstrating the effectiveness of the task framework proposed by READS. Secondly, the subgraph search task showed the greatest performance difference before and after ablation, indicating that the model's subgraph search capability is the most critical within the current framework. Lastly, entity type information has been proven to effectively assist large models in conducting more precise reasoning processes.

\subsection{Error Type Analysis}

\begin{figure}[ht]
    \centering
    \includegraphics[width=0.85\linewidth]{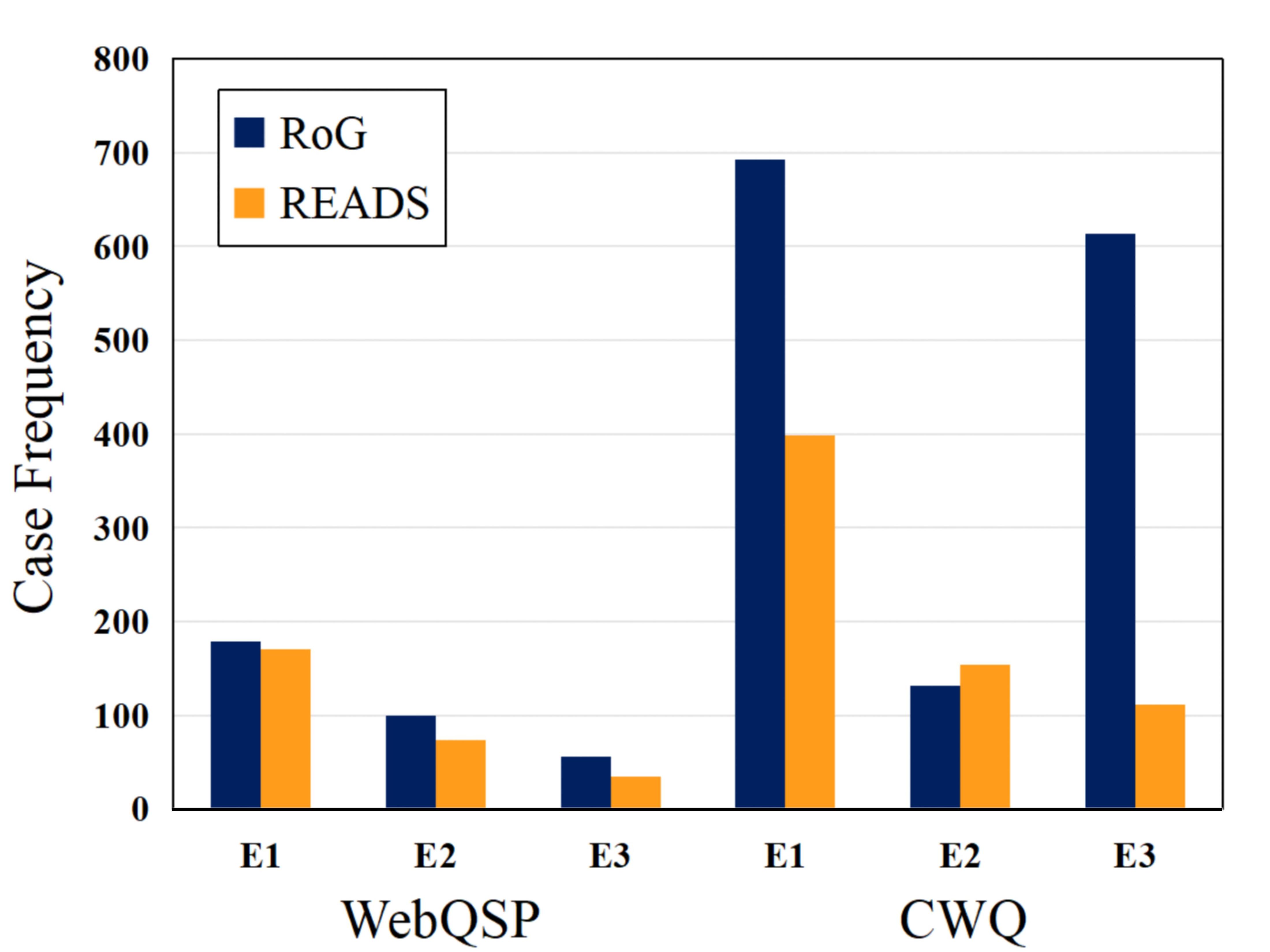}
    \caption{Case frequency of different types of errors, E1 corresponds to searching subtask; E2 corresponds to pruning subtask; E3 corresponds to answering subtask.}
    \label{fig:fig3}
\end{figure}

To analyze the effect of adopting READS on ungrounded reasoning with hallucinatory behavior, we collected and examined the frequency of error cases in READS. Since we decompose the KGQA process into three subtasks executed sequentially, we can categorize all errors into the following three types: 1) E1 stands for abscense of answer in the retrieved subgraph, corresponding to searching subtask; 2) E2 stands for lack of filering of the answer set, corresponding to pruning subtask; 3) E3 stands for mistakenly chosen the position of answer, corresponding to answering subtask (refer to Appendix~\ref{sec:appendix_error} for more details). We analyzed the frequency of these different error types, and the results are shown in Figure~\ref{fig:fig3}.

According to the result, compared to the generation-based method, READS significantly reduces the frequency of E3 errors on both benchmarks and also reduces the overall frequency of E2 errors, which proves that READS alleviates ungrounded reasoning behaviors in most cases by applying discriminative reasoning strategies. The dropped case frequency of E1 is consistent with our claim that the READS enhances subgraph searching capability of the LLM.

\subsection{Hallucination vs. Internal Knowledge}
\label{sec:gpt4}

As mentioned earlier, to mitigate hallucinations, READS abandons the capability of LLMs to generate answers directly, focusing instead on discriminative subtasks. To further analyze the balance between hallucination and LLM internal knowledge, we conduct experiments with a strong model GPT-4 and our base model llama2-7b using different answering strategies.

We employed three strategies: 1) Zero-shot: directly listing answers based on the question; 2) READS: using our proposed framework; 3) Augmented: generating answers based on subgraphs extracted by READS. Since we could not fine-tune or constrain the generation process of GPT-4, we presented it with a pool of options and asked it to make a selection. 

\begin{table}[h]
\centering
\scalebox{0.90}{
\renewcommand{\arraystretch}{1.1}
\begin{tabular}{l|cc|cc}
\hline
\multirow{2}{*}{\textbf{Strategy}} & \multicolumn{2}{c|}{\textbf{WebQSP}} & \multicolumn{2}{c}{\textbf{CWQ}} \\ \cline{2-5} 
                                     & GPT-4            & Llama2               & GPT-4          & Llama2             \\ \hline
Zero-shot                            & \textbf{0.632}             & 0.403            & \textbf{0.483}           & 0.297          \\ \hline
READS                            & 0.544             & \textbf{0.840}            & 0.346           & \textbf{0.802}          \\ \hline
Augmented                           & \textbf{0.856}             & 0.791            & \textbf{0.792}           & 0.632          \\ \hline
\end{tabular}
}
\caption{The Hits@1 under different strategies.}
\label{table:gpt4}
\end{table}

As shown in Table~\ref{table:gpt4}, compared to the READS process, GPT-4 achieves better results than Llama2 in generating answers based on subgraphs. We believe this illustrates the differences in strategic adaptability between scaled models and 7b-size models. For the commonly used 7b-size models, applying a constrained generation framework may better enhance their ability to inference answers.

\subsection{Subtask Data Efficiency}
We examine the training efficiency of the READS subtasks. We combine subgraph pruning and answer inference in the figure as reasoning component and fine-tune two separate models from scratch using Llama2-7b. When evaluating the performance of one model, we use the other model in its fully fine-tuned form.

\begin{figure}[h]
    \centering
    \begin{subfigure}[b]{0.47\columnwidth}
        \centering
        \begin{tikzpicture}
            \begin{axis}[
                width=4.4cm,  
                height=4.4cm, 
                legend pos=south east,
                ymajorgrids=true,
                grid style=dashed,
                xlabel={proportion of training data (\%)},
                xticklabels={$0$, $10$,$25$, $50$, $75$, $100$},
                ylabel={Hits@1},
		      xtick={0,1,2,3,4,5},
                tick label style={font=\tiny},
                xlabel style={font=\scriptsize,yshift=8pt},
                ylabel style={font=\scriptsize,yshift=-12pt},
            ]

            \addplot+[sharp plot, mark=square*,mark size=1.2pt,mark options={solid,mark color=red}, color=red] 
		coordinates
		{(0,0.786) (1,0.821) (2,0.822) (3,0.829) (4,0.826) (5,0.84)};
        \addlegendentry{\scriptsize{Reason}}
            \addplot+[sharp plot, mark=triangle*,mark size=1.2pt,mark options={solid,mark color=blue}, color=blue] 
		coordinates
		{(0,0.058) (1,0.798) (2,0.808) (3,0.828) (4,0.826) (5,0.84)};
        \addlegendentry{\scriptsize{Search}}
            \end{axis}
        \end{tikzpicture}
        \caption{WebQSP}
        \label{fig:sub-first3}
    \end{subfigure}
    \hspace{0.02\columnwidth}
    \begin{subfigure}[b]{0.47\columnwidth}
        \centering
        \begin{tikzpicture}
            \begin{axis}[
                width=4.4cm,  
                height=4.4cm,        
                legend pos=south east,
                ymajorgrids=true,
                grid style=dashed,
                xlabel={proportion of training data (\%)},
                xticklabels={$0$, $10$,$25$, $50$, $75$, $100$},
		      xtick={0,1,2,3,4,5},
                tick label style={font=\tiny},
                xlabel style={font=\scriptsize,yshift=8pt},  
                ylabel style={font=\tiny},
            ]

            \addplot+[sharp plot, mark=square*,mark size=1.2pt,mark options={solid,mark color=red}, color=red] 
		coordinates
		{(0,0.662) (1,0.801) (2,0.813) (3,0.809) (4,0.798) (5,0.809)};
        \addlegendentry{\scriptsize{Reason}}
            \addplot+[sharp plot, mark=triangle*,mark size=1.2pt,mark options={solid,mark color=blue}, color=blue] 
		coordinates
		{(0,0.14) (1,0.506) (2,0.773) (3,0.793) (4,0.806) (5,0.809)};
        \addlegendentry{\scriptsize{Search}}

            \end{axis}
        \end{tikzpicture}
        \caption{CWQ}
        \label{fig:sub-first4}
    \end{subfigure}
    \caption{The Hits@1 performance using different proportion of finetuning data.}
    \label{fig:data_efficiency}
\end{figure}

As shown in Figure~\ref{fig:data_efficiency}, both models require only about 25\% of the training data to reach near the best performances. In terms of data requirements among subgraph retrieval, subgraph pruning and answer inference, subgraph retrieval demands more data and poses greater challenges to the LLM.
% \begin{table}
% \centering
% \begin{tabular}{c|cc}
% \hline
%     & WebQSP & CWQ  \\ \hline
% ToG & 11.2   & 14.3 \\ \hline
% READS & 3.9    & 5.7  \\ \hline
% \end{tabular}
% \caption{The average model calls and input/output tokens required to answer a question.}
% \label{table:efficiency}
% \end{table}

\subsection{Further Analysis}

\textbf{Model Universality of READS.} To analyze the model Universality of READS, we test the performance of the READS method based on different backbone models, and the results are shown in Table~\ref{table:universality}. The results indicate that changing the base model has no significant impact on the method's performance, highlighting the universality of the READS approach.

\begin{table}[h]
\centering
\scalebox{0.90}{
\renewcommand{\arraystretch}{1.05}
\begin{tabular}{l|cc|cc}
\hline
\multirow{2}{*}{\textbf{Base Model}} & \multicolumn{2}{c|}{\textbf{WebQSP}} & \multicolumn{2}{c}{\textbf{CWQ}} \\ \cline{2-5} 
                                     & Hits@1            & F1               & Hits@1          & F1             \\ \hline
Vicuna-7b                            & 0.809             & 0.828            & 0.778           & 0.794          \\ \hline
Llama-7b                            & 0.830             & 0.842            & 0.799           & 0.823          \\ \hline
Llama2-7b                            & 0.840             & 0.845            & 0.802           & 0.820          \\ \hline
Llama3-8b                            & 0.827                   &0.845                  & 0.812           & 0.831          \\ \hline
Qwen2.5-7b                           &0.825                   &0.840                  &0.809                 &0.821                \\ \hline
\end{tabular}
}
\caption{Model universality of READS.}
\label{table:universality}
\end{table}

\noindent\textbf{Reasoning Cost.}
The interactive analysis between LLMs and knowledge graphs can be quite time-consuming, particularly when large subgraphs introduce long contexts that further hinder reasoning efficiency. However, through the implementation of a highly efficient reasoning strategy, READS has significantly reduced both the average number of model calls per question and the number of tokens per request. As demonstrated in Table~\ref{table:efficiency}, READS has halved the cost and achieved a similar average number of model calls as RoG, which plans a subgraph through a single generation.

\begin{table}[h]
\centering
\scalebox{0.9}{
\renewcommand{\arraystretch}{1.05}
\setlength{\tabcolsep}{1mm}
\begin{tabular}{l|ccc|ccc}
\hline
\multirow{2}{*}{\textbf{Method}} & \multicolumn{3}{c|}{\textbf{WebQSP}} & \multicolumn{3}{c}{\textbf{CWQ}} \\ \cline{2-7} 
                                     & input            & output              & calls         & input            & output               & calls             \\ \hline
RoG                            & 343.3             & 47.4            & 4.0           & 490.1           & 42.9             & 4.0\\ \hline
ToG                            & -             & -            & 11.2           & -           & -             & 14.3\\ \hline
READS                            & 178.9             & 10.8            & 3.9           & 206.6           & 12.4             & 5.7\\ \hline
\end{tabular}
}
\caption{The average model calls per question and average number of input/output tokens per request.}
\label{table:efficiency}
\end{table}

\noindent\textbf{Case Study.} We present cases of solving KGQA problems using the READS method in Appendix~\ref{sec:case}. READS provides effective explicit intermediate reasoning information, which adds to the readability of the overall KGQA process.

\section{Conclusion}
In this paper, we propose a novel LLM-based reasoning framework READS to reformulate KGQA process, aiming to alleviate the hallucination issues in existing generative methods and enhance the LLM's reasoning capability. Experimental results proves our claim that by decomposing KGQA and adopting designed discriminative strategies, we can enhances the capability of LLMs to retrieve question-related subgraphs and mitigate ungrounded reasoning results caused by hallucinations in the generation process.

\section*{Limitations}

Though our proposed READS framework has shown competitive KGQA performance and is proven to enhance the LLM's reasoning capability, we identify several limitations that requires further improvement. In the future, we will focus on the following directions to extend the current work:

1) Entity linking: Existing methods assume that the entity linking process is done before the KGQA process~\cite{rog,ToG}; In this work we follow the previous works to assume that the entity linking has already been completed. This is a common issue faced by the KGQA methods, we will explore how to eliminate this assumption to achieve reliable KGQA process.

2) Demand on labeled data: Although our method effectively enhances the knowledge reasoning capabilities of large models and demonstrates competitive performance across multiple datasets, we assume the existence of a gold query. Given the strong zero-shot KGQA capability and reasoning capability of GPT-4, works that does not rely on a gold query either requires GPT-4 to annotate the reasoning process (such as RoG) or combines the knowledge memory of strong models to improve overall performance (such as ToG, GoG, etc.)~\cite{rog,ToG,xu2024generate}. In the future works, we will explore the possibility of using model-generated pseudo-labels or constructing self-summarized memories to deal with this issue.

% 3) Implementation with scaled models: Although we got competitive performance among widely used LLMs with 7b parameters, we found it technically challenging to implement discriminative subtasks on large models such as GPT-4 or LLaMA2-70B. On one hand, for API models such as GPT-4, it is hard to implement constrained generation process to facilitate the discriminative subtasks, while demonstrating the available choices will significantly increase the interaction cost (on average 70 options per model call) and may result in unexpected errors. On the other hand, for large models like LLaMA2-70B, directly employing constrained generation processes can easily result in out of memory issues. In the future, we will explore the possibility of scaled model's  techniques remains to be integrating discriminative generation processes with these scaled and strong models remains to be 
% is a research direction that warrants exploration.

% Bibliography entries for the entire Anthology, followed by custom entries
%\bibliography{anthology,custom}
% Custom bibliography entries only
\bibliography{custom}

\newpage

\appendix

\section{Semantic Entity Types}
\label{sec:appendix_A}

Here we demonstrate different semantic entity types in Table~\ref{table:ent_type}.

\begin{table}[]
\centering
\begin{tabular}{c|cc}
\hline
\textbf{Type} & \textbf{Definition}                                                                                                                                    & \textbf{Example}                         \\ \hline
entity               & \begin{tabular}[c]{@{}c@{}}Real entities include\\ person\textbackslash{}school\textbackslash{}events\\ and so on\end{tabular}                         & Micheal                                  \\ \hline
topic                & \begin{tabular}[c]{@{}c@{}}Topic id entities which\\ is used to connect\\ entities with the\\ same topic, its id\\ has no actual meanings\end{tabular} & m.01428y                               \\ \hline
num                  & Numbers                                                                                                                                                & 240.15                                   \\ \hline
date                 & Dates                                                                                                                                                  & 2015\textbackslash{}08\textbackslash{}10 \\ \hline
\end{tabular}
\caption{Entity types with its definition and example}
\label{table:ent_type}
\end{table}

\section{Data Preprocessing}
\label{sec:appendix_B}

\textbf{Training Data for Subgraph Searching.} We make use of the SPARQL data available in existing benchmarks to form the training data. In WebQSP and CWQ, each question is associated with a SPARQL query. The direct execution of this query yields the answer to the open question. We obtain the correct subgraph structure required to solve each problem by decomposing the SPARQL statements. Unlike ROG (\citealp{rog}), in finetuning process READS always presents the model with the correct knowledge subgraph structure rather than the shortest path starts from the question entity and ends at the answer entities.

\noindent\textbf{Training Data for Subgraph pruning and answer inference.} To finetune the LLMs to be capable of constraint determination and answer inference, we also construct constraint/answer locating samples from the SPARQL queries in WebQSP and CWQ. The input is a complete subgraph structure with all feasible options of constraints or answer positions, the golden output is the correct position of the constraint and the answer.

\noindent\textbf{Freebase preprocessing.} Due to the huge volume of established Freebase knowledge graph, directly interacting with Freebase through SPARQL is inefficient and may result in unnecessary syntax errors. Following UniKGQA (\citealp{jiang2022unikgqa}), we extract subgraphs from Freebase using breadth-first search for each question, which are then used for the subgraph searching process. Additionally, we expand these subgraphs using the SPARQL queries provided in the benchmarks to ensure the presence of constraint branches.

SPARQL queries contain the subgraph information necessary to complete a comprehensive graph query. These queries are composed of graph structure triples and filtering conditions. In WebQSP, CWQ, and most KGQA datasets built on Freebase, each question corresponds to a specific SPARQL query. Therefore, the paths included in SPARQL effectively represent the correct subgraph structure required to answer the current question.

In previous works, subgraphs obtained using shortest path search methods typically formed chain-like structures. Compared to the information contained in SPARQL, these structures: 1) might not be logically coherent search paths, and 2) could miss some branches on certain nodes along the path. To enable our method to proceed smoothly, we extracted additional subgraph structures with all possible branches related to the question from Freebase based on the structural information inherent in the SPARQL queries. These were added to the original dataset (for a reference to the original dataset, see RoG). The specific implementation can be found in the corresponding functions in the open-source code, and will not be elaborated here. 

\section{Metrics}
\label{sec:appendix_C}

Here we outline the metrics calculation formulas and their corresponding meanings that were not detailed in the main text. 

\noindent\textbf{Hits@1.} Hits@1 calculates the proportion of questions for which the first answer given by the model is correct. Given $A_{pre}$ is the predicted list of answers, and $A_{gold}$ is the list of golden answers, $A_{pre}[0]$ as the very first answer the model predict, then we have:
\begin{equation}
    Hits@1=\frac{count(A_{pre}[0]\in A_{gold})}{count(questions)}.
\end{equation}

For example, if the correct answer is "apple" and the model answers "pear, apple, banana," then Hits@1 for this question is 0. It is important to note that this metric can sometimes be miscalculated as follows:
\begin{equation}
    Hits@1=\frac{count(A_{pre}\cap A_{gold} \neq \emptyset)}{count(questions)}.
\end{equation}

With this incorrect calculation, the Hits@1 would be higher. For the above example, the Hits@1 for this question would be 1.

\noindent\textbf{F1.} We adopt the same calculation method as previous work, using the Macro-F1 scoring method. First, we calculate the precision and recall for each test sample. Then, we average them based on the number of samples to obtain the overall recall and precision. Finally, we use the harmonic mean of the overall recall and precision to calculate the overall F1 score.

\begin{table}[]
\scalebox{0.85}{
\renewcommand{\arraystretch}{1.5}
\centering
\begin{tabular}{c|cccc}
\toprule[1pt]
Leaf Node Number & 2     & 3     & 4     & 5     \\ \hline
with threshold    & 0.847 & 0.741 & 0.625 & 0.604 \\ \hline
w/o threshold   & 0.845 & 0.750 & 0.636 & 0.738 \\ \bottomrule[1pt]
\end{tabular}
}
\caption{The average Hits@1 performance on questions with different subgraph structures. Manually add minimum branch threshold during tree search process. The performance drops as we manually add the threshold.}
\label{table:thres}
\end{table}

\begin{table}[]
\centering
\scalebox{0.95}{
\setlength{\tabcolsep}{1mm}
\begin{tabular}{c|cccc}
\hline
\multirow{2}{*}{\textbf{Method}} & \multicolumn{4}{c}{\textbf{GrailQA Dev}} \\ \cline{2-5} 
                                     & i.i.d            & compositional              & zero-shot         & overall \\ \hline
PANGU                            & 0.844             & 0.746            & 0.716           & 0.754           \\ \hline
READS                            & 0.921             & 0.759            & 0.626           & 0.718          \\ \hline
\end{tabular}
}
\caption{The Hits@1 performance on GrailQA.}
\label{table:grailqa}
\end{table}

\section{Ungrounded Reasoning Behavior}
\label{sec:appendix_D}
The previous generation-based method can sometimes provide the correct answer even when the subgraph does not contain the correct answer, whereas READS does not exhibit this behavior (see case C2 in Table~\ref{table:case_count}).

\section{Result on GrailQA}
\label{sec:grailqa}

We test our proposed READ on more challenging benchmark GrailQA's development set~\cite{gu2021beyond}, the results are shown in Table~\ref{table:grailqa}. Compared to the previous single task discriminative method PANGU~\cite{gu2023don}, although not achieving overall SOTA performance, READS enhanced the KGQA reliability on both i.i.d and compositional questions, which proves the effectiveness of the reformulation strategy used in READS.

\section{Types of Errors}
\label{sec:appendix_error}

\begin{table*}[]
\centering
\scalebox{0.85}{
\renewcommand{\arraystretch}{1.5}
\begin{tabular}{c|c|c}
\hline
\textbf{\begin{tabular}[c]{@{}c@{}}Does retrieved\\ subgraph contains correct answer?\end{tabular}} & \textbf{Is the very first answer predicted correct?}                & \textbf{Case type} \\ \hline
\multirow{3}{*}{Yes}                                                                                & Yes                                                      & C1                 \\ \cline{2-3}
                                                                                                    & No, but the correct answer exist in the predicted list   & E2                 \\ \cline{2-3} 
                                                                                                    & No, and there is no correct answer in the predicted list & E3                 \\ \hline
\multirow{2}{*}{No}                                                                                 & Yes                                                      & C2                 \\ \cline{2-3} 
                                                                                                    & No                                                       & E1                 \\ \hline
\end{tabular}
}
\caption{Error Case type definitions.}
\label{table:case_def}
\end{table*}

\begin{table*}[]
\centering
\scalebox{0.9}{
\renewcommand{\arraystretch}{1.5}
\begin{tabular}{lclcl}
\hline
\multicolumn{1}{l|}{\textbf{Case Type}}                                                                                                                             & \multicolumn{2}{l|}{\textbf{CWQ}}                                           & \multicolumn{2}{l}{\textbf{WebQSP}}                    \\ \hline
\multicolumn{1}{l|}{\textbf{RoG}}                                                                                                                                   & \multicolumn{1}{l}{\textbf{Total}} & \multicolumn{1}{l|}{\textbf{Seperate}} & \multicolumn{1}{l}{\textbf{Total}} & \textbf{Seperate} \\ \hline
\multicolumn{1}{l|}{{\color[HTML]{333333} C1}}                                                                             &                                & \multicolumn{1}{l|}{1645}              &                               & 1208              \\ \cline{1-1}
\multicolumn{1}{l|}{{\color[HTML]{333333} \begin{tabular}[c]{@{}l@{}}E2\end{tabular}}}  &   2390                                 & \multicolumn{1}{l|}{132}               & 1363                                   & 99                \\ \cline{1-1}
\multicolumn{1}{l|}{{\color[HTML]{333333} E3}}                                                                               &                                    & \multicolumn{1}{l|}{613}               &                                    & 56                \\ \hline
\multicolumn{1}{l|}{{\color[HTML]{333333} C2}}                                                                    &                                    & \multicolumn{1}{l|}{{\ul 364}}         &                                    & {\ul 78}          \\ \cline{1-1}
\multicolumn{1}{l|}{{\color[HTML]{333333} E1}}                                                                      & \multirow{-2}{*}{1057}             & \multicolumn{1}{l|}{693}               & \multirow{-2}{*}{257}              & 179               \\ \hline
\multicolumn{5}{l}{\textbf{READS}}                                                                                                                                                                                                                                                                        \\ \hline
\multicolumn{1}{l|}{{\color[HTML]{333333} C1}}                                                                             &                                    & \multicolumn{1}{l|}{2784}              &                                    & 1342              \\ \cline{1-1}
\multicolumn{1}{l|}{{\color[HTML]{333333} \begin{tabular}[c]{@{}l@{}}E2\end{tabular}}} &                                    & \multicolumn{1}{l|}{154}               &                                    & 73                \\ \cline{1-1}
\multicolumn{1}{l|}{{\color[HTML]{333333} E3}}                                                                               & \multirow{-3}{*}{3049}             & \multicolumn{1}{l|}{111}               & \multirow{-3}{*}{1449}             & 34                \\ \hline
\multicolumn{1}{l|}{{\color[HTML]{333333} C2}}                                                                    &                                    & \multicolumn{1}{l|}{{\ul 0}}           &                                    & {\ul 0}           \\ \cline{1-1}
\multicolumn{1}{l|}{{\color[HTML]{333333} E1}}                                                                      & \multirow{-2}{*}{398}              & \multicolumn{1}{l|}{398}               & \multirow{-2}{*}{171}              & 171               \\ \hline
\end{tabular}
}
\caption{Frequency count of different cases.}
\label{table:case_count}
\end{table*}

\begin{table*}[]
\centering
\scalebox{1}{
\renewcommand{\arraystretch}{1.2}
\begin{tabular}{c|cccccccc|c|c}
\hline
\multirow{2}{*}{\textbf{Leaf Count}} & \multicolumn{8}{c|}{\textbf{Edge count in the subgraph}}                                             & \multirow{2}{*}{\textbf{Total}} & \multirow{2}{*}{\textbf{Percentage}} \\ \cline{2-9}
                                     & \textbf{1} & \textbf{2} & \textbf{3} & \textbf{4} & \textbf{5} & \textbf{6} & \textbf{7} & \textbf{8} &                                 &                                      \\ \hline
2                                    & 921        & 1453       & 1267       & 400        & 22         & 0          & 0          & 0          & 4063                            & 0.802                                \\
3                                    & 0          & 0          & 278        & 217        & 208        & 5          & 0          & 0          & 708                             & 0.139                                \\
4                                    & 0          & 0          & 0          & 71         & 41         & 44         & 2          & 3          & 161                             & 0.031                                \\
5                                    & 0          & 0          & 0          & 0          & 34         & 40         & 57         & 0          & 131                             & 0.025                                \\ \hline
\end{tabular}
}
\caption{Statistics of questions with different knowledge subgraph structure. This is the statistic result combining WebQSP and CWQ's test set.}
\label{table:sta_subgraph}
\end{table*}

\begin{table*}
\resizebox{1\linewidth}{!}{
\begin{tabular}{l}
\hline
\textbf{Tree Search Stage Prompt Template}\\
\hline
Below is a wikipedia question,\\
you can retrieve a graph to help you answer the question.\\
The retrieved graph information is given as information triple like (Entity1, Relation, Entity2),\\
or only the name of the start entity. \\
Decide which entity and corresponding relation to retrieve next,\\
response in form of 'entity+relation'. \\
Response 'None' if the retrieved graph is\\
informative enough to answer the question.\\
Question:\\
<The Question>\\
Retrieved graph:
<The retrieved subgraph structure $S_{k}$, in forms of triples>\\
Next retrieve:\\
\hline
\textbf{Tree Pruning Stage Prompt Template}\\
\hline
\textbf{Locate Constraints}:\\
Below is a question with a support graph presented as\\ triples (entity A, relation, entity B). \\
The entity name in the support graph is 'type\_id'. \\
'Type' denotes the entity type, which includes four types:\\ ordinary entity (entity), topic entity (topic),\\
number (num), and date (date). \\
'Id' is an incremental identifier used to distinguish entities. \\
Please match all the constrains with one of the entity in the support graph.\\
Support graph:\\
<The retrieved subgraph structure $S_{k}$, in forms of triples>\\
Question:\\
<The Question>\\
Constraints:\\
<List of constraint entities>\\
Determine result:\\ \hline
\textbf{Locate Answer}:\\
Below is a question with a support graph presented as\\ triples (entity A, relation, entity B). \\
The entity name in the support graph is 'type\_id'. \\
'Type' denotes the entity type, which includes four types:\\ ordinary entity (entity), topic entity (topic),\\
number (num), and date (date). \\
'Id' is an incremental identifier used to distinguish entities. \\
Please select the answer from the support graph by choosing the right entity.\\
Support graph:\\
<The retrieved subgraph structure $S_{k}$, in forms of triples>\\
Question:\\
<The Question>\\
Answer entity:\\ 
\hline
\end{tabular}
}
\caption{Prompt Template use in READS}
\label{table:prompt}
\end{table*}

We categorize the answering process into five scenarios, with three of these ultimately resulting in incorrect answers. We present the overall definitions in Table~\ref{table:case_def} and the frequency statistics for the five scenarios in Table~\ref{table:case_def}. Here we further explain the definition of the three error cases.

E1 (failed subgraph searching) is directly related to the graph search ability of the model. If the answer is not included in the retrieved subgraph, the model can not actually obtain the answer through inference and pruning.

We attribute E2 to lack of subgraph pruning as it indicates the presence of incorrect answer entities at the selected answer position. We detect E2 as the cases when the first answer is wrong all the correct answers are listed after the  
The lack of pruning may be caused by: 1) Omission of branches in the structure, which means the LLM fails to retrieve necessary entities; 2) Failure on matching constraints with the correct position. To avoid such errors, the model should have stronger searching and constraint locating capabilities.

We attribute E3 to wrong answer location since the answer list contains no golden answer. Although generation-based methods generates the answer rather than selecting the position, the inability to infer the answer from the graph containing the correct answer is considered as a similar location error. To avoid such errors, the model should have stronger subgraph reasoning and answer positioning capabilities.

\section{Statistics of Subgraph Structure}
\label{appendix:sta_subgraph}

After categorizing questions based on the structure of their corresponding knowledge subgraphs, we count the number of questions in each class(see Figure~\ref{table:sta_subgraph}), and find that there is a relative scarcity of graph-structured data with single or multiple branches.

Many questions with leaf count 2 is free from constraints, while these issues make up the vast majority(80.2\%) of the test set. This proportional relationship also appears in the training set, which means the model will see more simple graph structures during training process. This may lead the model to prematurely halt the search by favoring structures with fewer branches. However, introducing minimum branching threshold to force the LLM to search more branches before it terminates the search stage may obstacle normal tree search behavior (see Table~\ref{table:thres}). This remains a topic worth to be discussed in the future.

\begin{table*}[]
\centering
\begin{tabular}{l}
\toprule[1pt]
\textbf{Case 1}   \\ \toprule[1pt]
\textbf{Question:} \\ what does jamaican people speak?\\
\textbf{Tree search stage output $S_k$:} \\
('Jamaica', 'location.country.languages\_spoken', 'entity\_1')\\
\textbf{Real subgraph $G_k$ retrieve from Freebase:}\\
('Jamaica', 'location.country.languages\_spoken', 'Jamaican English')\\
('Jamaica', 'location.country.languages\_spoken', 'Jamaican Creole English Language')\\
\textbf{Tree pruning stage output:}\\
No constrain, Answer is “entity\_1”\\
\textbf{READS Output:}\\
'Jamaican English', 'Jamaican Creole English Language' \\ \toprule[1pt]
\multicolumn{1}{l}{\textbf{Case 2}} \\ \toprule[1pt]
\textbf{Question:} \\ who plays the voice of kitt in knight rider?\\
\textbf{Tree search stage output $S_k$:} \\
("Knight Rider", "tv.tv\_program.regular\_cast", "topic\_1")\\
("topic\_1", "tv.regular\_tv\_appearance.character", "entity\_2")\\ 
("topic\_1", "tv.regular\_tv\_appearance.actor", "entity\_3")\\
\textbf{Real subgraph $G_k$ retrieve from Freebase:}\\
("Knight Rider", "tv.tv\_program.regular\_cast", "m.03lj4m5")\\
("m.03lj4m5", "tv.regular\_tv\_appearance.character", "KITT")\\
("m.03lj4m5", "tv.regular\_tv\_appearance.actor", "William Daniels")\\
...\\
("m.02h9cb0", "tv.regular\_tv\_appearance.actor", "David Hasselhoff")\\
("m.03lj4m0", "tv.regular\_tv\_appearance.actor", "Edward Mulhare")\\
("m.03lj4lk", "tv.regular\_tv\_appearance.actor", "Patricia McPherson")\\
\textbf{Tree pruning stage output:}\\
Constraint: "entity\_2= KITT", Answer is “entity\_3”\\
\textbf{READS Output:}\\
'William Daniels' \\ \toprule[1pt]                 \textbf{Case 3}                     \\ \toprule[1pt]
\textbf{Question:} \\ where did kevin love go to college?\\
\textbf{Tree search stage output $S_k$:} \\
("Kevin Love", "people.person.education", "topic\_1")\\
("topic\_1", "education.education.institution", "entity\_2")\\
("entity\_2", "common.topic.notable\_types", "entity\_3")\\
\textbf{Real subgraph $G_k$ retrieve from Freebase:}\\
("Kevin Love", "people.person.education", "m.04ftwdc")\\("Kevin Love", "people.person.education", "m.045z4gx")\\ ("m.04ftwdc", "education.education.institution", "University of California, Los Angeles")\\
("m.045z4gx", "education.education.institution", "Lake Oswego High School")\\
("University of California, Los Angeles", "common.topic.notable\_types", "College/University")\\
("Lake Oswego High School", "common.topic.notable\_types", "School")\\
\textbf{Tree pruning stage output:}\\
Constraint: "entity\_3 = College/University", Answer is “entity\_2”\\
\textbf{READS Output:}\\
'University of California, Los Angeles' \\  \bottomrule[1pt]
\end{tabular}
\caption{Case Study of READS}
\label{table:case}
\end{table*}

\section{Prompt Templates}
\label{appendix:prompt}

We demonstrate all the prompt templates used in READS in Table~\ref{table:prompt}, including the template for tree searching, locating constraints and the answer.

\section{Case Study}
\label{sec:case}

We present two clear process examples of conducting KGQA tasks using READS in Table~\ref{table:case}.

\end{document}